# REHEARSE-3D: A Multi-modal Emulated Rain Dataset for 3D Point Cloud De-raining

Abu Mohammed Raisuddin[1], Jesper Holmblad[1], Hamed Haghighi[2], Yuri Poledna[3], Maikol Funk Drechsler[3], Valentina Donzella[2], and Eren Erdal Aksoy[1]

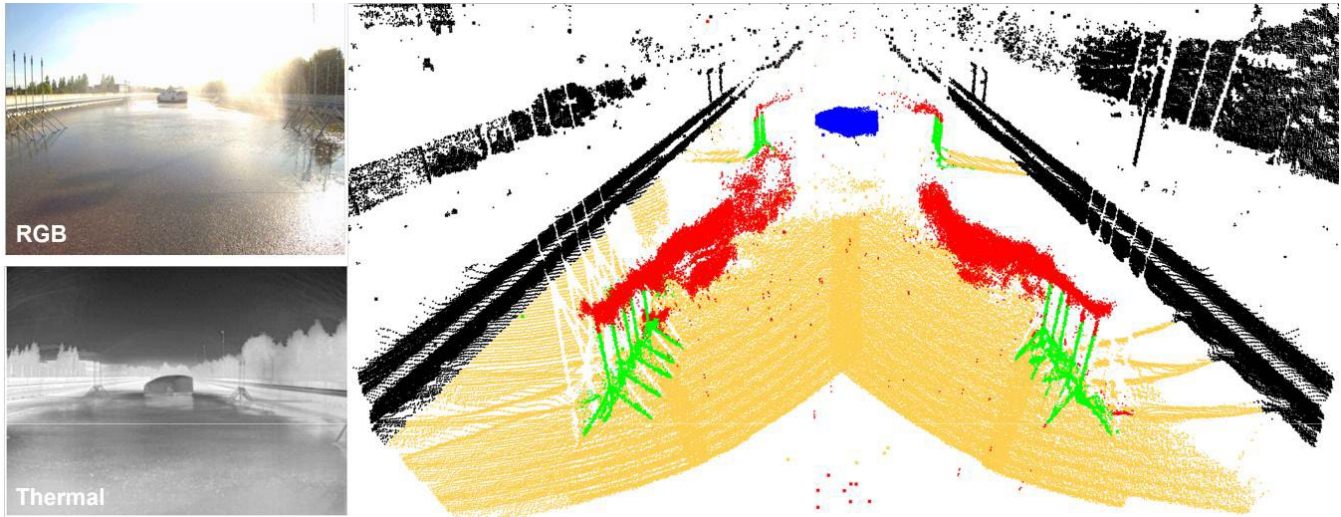

Fig. 1: A sample scene from REHEARSE-3D. Fully annotated high-resolution LiDAR and fused 4D Radar point clouds are shown on the right. Each color represents a unique semantic class: rain, car, sprinkler, pedestrian, biker, road, targets, and background. For visualization purposes only, the corresponding RGB and thermal camera images are shown on the left.

*Abstract*— Sensor degradation poses a significant challenge in autonomous driving. During heavy rainfall, the interference from raindrops can adversely affect the quality of LiDAR point clouds, resulting in, for instance, inaccurate point measurements. This, in turn, can potentially lead to safety concerns if autonomous driving systems are not weather-aware, i.e., if they are unable to discern such changes. In this study, we release a new, large-scale, multi-modal emulated rain dataset, REHEARSE-3D, to promote research advancements in 3D point cloud de-raining. Distinct from the most relevant competitors, our dataset is unique in several respects. First, it is the largest point-wise annotated, and second, it is the only one with high-resolution LiDAR data (LiDAR-256) enriched with 4D Radar point clouds logged in both daytime and nighttime conditions in a controlled weather environment. Furthermore, REHEARSE-3D involves rain-characteristic information, which is of significant value not only for sensor noise modeling but also for analyzing the impact of weather at a point level. Leveraging REHEARSE-3D, we benchmark raindrop detection and removal in fused LiDAR and 4D Radar point clouds. Our comprehensive study further evaluates the performance of various statistical and deep-learning models. The dataset and benchmark models are publicly available at: https://sporsho.github.io/REHEARSE3D.

## I. INTRODUCTION

Autonomous driving technology has undergone significant advancements in recent years. However, the reliability and performance of autonomous vehicles in inclement weather conditions remain a significant challenge. Among these challenges, rainfall poses a particularly complex problem, as raindrops can adversely impact the performance of sensors, thus hindering downstream perception tasks [1].

LiDAR sensors have become a staple in autonomous vehicles due to their high accuracy, reliable depth calculation, and long-range measurements. However, the integrity of these estimations can significantly deteriorate due to precipitation, such as raindrops, which can scatter and refract the laser beams. Precipitation characteristics, such as intensity and frequency, exert a substantial influence on the accuracy of the resulting point clouds. For instance, heavy rainfall can substantially degrade the quality of the 3D LiDAR point cloud, causing inaccurate and sparser point clouds. This, in turn, can lead to miss-detection and false positive predictions in downstream perception tasks [2–4].

To address these weather-related challenges, numerous

*Co-funded by the European Union. Views and opinions expressed are however those of the author(s) only and do not necessarily reflect those of the European Union or European Climate, Infrastructure and Environment Executive Agency (CINEA). Neither the European Union nor the granting authority can be held responsible for them. Project grant no. 101069576.

[1]Halmstad University, School of Information Technology, Cen-ter for Applied Intelligent Systems Research, Halmstad, Sweden abu-mohammed.raisuddin@hh.se

[2]WMG, University of Warwick, Coventry, United Kingdom

[3]CARISSMA Institute of Automated Driving, Technische Hochschule Ingolstadt, Ingolstadt, Germany

studies have been conducted not only by introducing diverse datasets to capture the impact of rainfall on sensor readings [5–7] but also by identifying and filtering out precipitation-related noisy measurements (a.k.a. de-raining), particularly those caused by raindrops in 3D point clouds [3, 6, 8]. However, existing annotated datasets are unimodal (using LiDAR only) with limited data density (max 64-beam scans) and lack diversity in rain intensity (see Table I). These limitations impede the efficacy of benchmarks for point-cloud de-raining.

To push the boundaries of research in this area, we present REHEARSE-3D, a novel, large-scale, and multi-modal em-ulated rain dataset, including high-resolution LiDAR-256 scans complemented with weather-resilient 4D Radar point clouds logged in both daytime and nighttime rainy condi-tions. REHEARSE-3D is enriched with semantic information for each LiDAR and Radar point (see Figure 1) to facilitate research in 3D point cloud de-raining for autonomous driving applications. Furthermore, REHEARSE-3D provides precip-itation characteristic information, including rain intensity, droplet size distribution, wind velocity and direction, and visibility. This information is of significant value not only for sensor noise modeling but also for analyzing the impact of weather at a point level. To the best of our knowl-edge, REHEARSE-3D is the first comprehensive multi-modal point-wise annotated high-density point cloud dataset capturing various rain effects across day and night conditions in a controlled weather environment.

REHEARSE-3D is the semantically enhanced iteration of the REHEARSE (adveRse wEatHEr datAset for sensoRy noiSe modEls) dataset [9] conducted in a controlled outdoor rain facility with target objects (e.g., pedestrians, cars, bikers, etc.) at fixed positions. For rain emission, REHEARSE employs a set of mobile water sprinklers to generate synthetic rain in three precipitation levels: light (10 mm/h), medium (25 mm/h), and heavy (50 mm/h). The dataset involves an automotive-grade sensor suite, including not only high-resolution LiDARs and a 4D Radar but also RGB and thermal cameras. We note that REHEARSE-3D provides point-wise semantic annotations only for MEMS LiDAR and 4D Radar readings.

Leveraging REHEARSE-3D, we establish a comprehen-sive benchmark for de-raining fused LiDAR and 4D Radar point clouds. We evaluate the performance of several state-of-the-art statistical and deep-learning baseline models to provide an in-depth analysis. The objective is to establish strong baseline performance standards that would facilitate and advance future research efforts in 3D point cloud de-raining.

In short, our contributions are listed below:
- We introduce a new large-scale multi-modal emulated rain dataset, named REHEARSE-3D, with 9.2 billion point annotations, logged in various rain intensities in daytime and nighttime conditions in a controlled weather environment.
- Leveraging REHEARSE-3D, we benchmark various state-of-the-art denoising algorithms to de-rain the early-fused LiDAR and Radar point clouds.
- We use the point cloud from clean weather and statisti-cally generate synthetic raindrops to study the emulated-to-simulated domain gap.

II. RELATED WORK

A substantial corpus of research has been devoted to multi-modal datasets for autonomous driving [10–14], as comprehensively outlined in [15, 16]. When it comes to adverse weather conditions, datasets are scarce since some concentrate on individual modalities, such as LiDAR [8, 17] or Radar [18], or address specific tasks, such as object de-tection and localization, as in the case of [19–21], or exclude annotations for outliers, such as raindrops or snowflakes, as in [17, 22], despite providing ground-truth point-wise seman-tic annotations for the full 3D point cloud data. Additionally, there exist synthetically corrupted datasets to simulate severe weather conditions [23–25], which inevitably suffer from the simulation-to-reality gap.

Regarding the emulated weather datasets, TWICE [28] introduced a synthetic rain dataset using a camera and a 3D Radar. The logged 2D rain data is, however, neither properly validated nor involves any annotations. The work in [7] solely addresses the degradation of LiDAR and camera readings in rainy conditions, conceptualizing weather moni-toring as a regression task, for instance, by estimating rainfall intensity. However, no semantic information is provided for the logged sparse 32-beam LiDAR data.

Only a few relevant studies provide detailed point-wise ground-truth annotations, even for outliers, such as rain-drops and/or snowflakes, in harsh weather data [5, 8, 27]. WADS [8] is a real 3D LiDAR point cloud dataset with pointwise labels for falling snow and accumulated snow. This dataset, however, lacks other sensor modalities and rainy conditions. Likewise, SnowyKITTI [27] is a simulated snowy iteration of the real SemanticKITTI [26] dataset coming with LiDAR-only sensor readings. SemanticSpray [5] is a real 3D point cloud dataset scanned with LiDAR-32. This dataset is based on RoadSpray [29]; however, it only contains scenes after rainfall of varying intensities. Therefore, the spray effect is generated by the ego vehicle and the individual vehicles surrounding the ego vehicle. WeatherNet [6] is the closest work to ours. It provides emulated rain and fog point cloud data logged in an indoor rain/fog chamber. This dataset exhibits a notable deficiency in sensor resolution (32-beam LiDAR), with ground-truth annotations encompassing only a forward-facing view of about $60°$ in a horizontal field, thus substantially diminishing the number of labeled points.

Most of these aforementioned datasets contribute to the field by providing 3D point-wise semantic annotations for sequential LiDAR scans. As shown in [2, 3], such annota-tions are crucial in detecting and removing noisy points to further enhance downstream perception tasks (e.g., semantic segmentation and object detection) under adverse weather conditions. However, all these datasets focus only on LiDAR data with limited resolutions (max 64-beam, as in the case of WADS [8]). LiDAR point clouds are well known for

TABLE I: Comparison of relevant outdoor datasets with REHEARSE-3D. *SnowyKITTI is generated from SemanticKITTI after applying the snow generation process to it. #points is in Millions. The SemanticSpray dataset does not include live rain; however, it does contain scenes following a rainfall in which the road exhibits three distinct levels of water accumulation. The precise quantitative data concerning the rainfall is not included in the dataset. Our dataset contains three levels (10mm, 25mm, and 50mm) of emulated rain scenes generated from sprinklers.

| | #Points | #Classes | Modality | Annotation | Sequential | Weather | Rain Characteristics | Day/Night | Environment |
|---|---|---|---|---|---|---|---|---|---|
| SemanticKITTI [26] | 4549 | 28 | LiDAR-64 | Point-wise | ✓ | Clean | - | - | Real |
| SnowyKITTI [27] | 3940 | 2 | LiDAR-64 | Point-wise | ✓ | Snow | - | - | Simulated |
| WADS [8] | 387 | 22 | LiDAR-64 | Point-wise | ✓ | Snow | - | - | Real |
| SemanticSpray [5] | 526 | 3 | LiDAR-32 | Point-wise | ✓ | Rain | - | - | Real |
| WeatherNet [6] | 1700 | 3 | LiDAR-32 | Point-wise | ✓ | Rain/Fog | ✓ | - | Real |
| REHEARSE-3D (Ours) | 9200 | 8 | LiDAR-256* & 4D-Radar | Point-wise | ✓ | Rain/Clean | ✓ | ✓ | Real |

*The LiDAR is a MEMS LiDAR with 256 lines.

being sparse, making it challenging to develop perception solutions for them. On top of that, inclement weather makes it even worse because the interaction between the laser beam and the rain drop/snow particle makes the point cloud even sparser. Such a drop in sparsity can be seen in the work of SemanticSpray [5] and WeatherNet [6]. This inevitable sparsity issue warrants the collection of a high-density point cloud.

Table I compares our dataset with these highly relevant counterparts. Our proposed REHEARSE-3D differs from all these existing competitors in that it is the largest annotated dataset and the only one with high-resolution LiDAR data (LiDAR-256) enriched with 4D Radar point cloud. We note that REHEARSE-3D also has RGB, thermal camera, and weather sensor readings, which are, however, not semantically labeled. Like WeatherNet [6], REHEARSE-3D additionally provides rain-characteristic information, which paves the way for grounding the computational modeling of emulated raindrops. This information is of significant value when comparing simulated and real counterparts, thereby facilitating the process of bridging the sim-to-real gap. Consequently, Table I conveys that REHEARSE-3D makes a significant contribution to the field by providing the largest point-wise annotated high-density point clouds that are logged in various rain intensities in both daytime and nighttime conditions in a controlled weather environment.

### III. THE REHEARSE-3D DATASET

#### A. Background

The REHEARSE (adveRse wEatHEr datAset for sensoRy noiSe modEls) dataset [9] is introduced to facilitate the enhancement, comparison, and benchmarking of automotive perception sensor models under "controlled" and "fully characterized" severe weather conditions, including rain and fog. As a weather-aware dataset designed explicitly for autonomous driving systems, REHEARSE makes a valuable contribution to the research community. REHEARSE enables the rehearsing and comparing sensor noise models in both simulated and real-world environments.

The REHEARSE dataset employs four automotive-grade sensors, including both passive and active automotive-grade sensors. Two distinct LiDAR technologies are incorporated: one featuring a rotating mechanical system (Ouster 128) and another utilizing micro-electromechanical systems (MEMS). A 4D Radar is included, which provides information on distance, velocity, azimuth, and elevation as its fourth dimension. The sensor suite is completed with an RGB camera and a FLIR thermal camera. Figure 2 illustrates the arrangement of the REHEARSE sensor suite.

REHEARSE utilizes a set of mobile water sprinklers to generate synthetic rain while regulating the rain intensity in the CARISSMA outdoor test track (See Figure 2). Three rain intensities are measured: 10 mm/h with a maximum span of 112 m, 25 mm/h with a maximum span of 56 m, and 50 mm/h with a maximum span of 28 m. These values are obtained during both nighttime and daytime testing. REHEARSE has a range of target/sensor distances, varying from 10 to 112 meters, and encompasses diverse weather conditions, including clear skies, heavy rainfall, and fog, with varying intensity levels. For additional technical details regarding the data logging setup, please refer to [30].

Additionally, REHEARSE offers critical parameters for describing weather phenomena, including rain intensity, droplet size distribution, wind velocity and direction, and visibility. This enhances the fidelity of sensor models. Weather effects, rain, and fog are validated using disdrometers, the square-meter method, weather stations, and visibility measurements. Weather conditions are evaluated using real rain data and the theoretical Marshall-Palmer droplet size distribution. This approach demonstrates that the generation

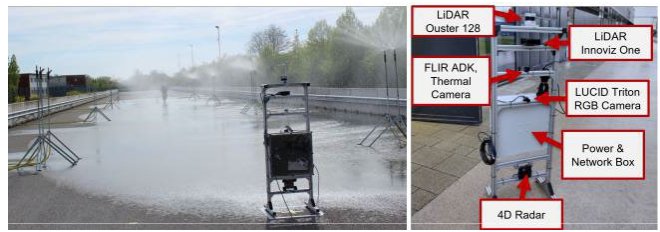

Fig. 2: CARISSMA Outdoor test track (left) and REHEARSE sensor setup (right).

of weather phenomena approximates real and theoretical counterparts.

### B. Data Annotation

REHEARSE-3D focuses on the clean and rainy REHEARSE data logged exclusively in the CARISSMA Out-door facilities. REHEARSE-3D is the semantically enhanced iteration of the original REHEARSE dataset [9], briefly introduced in Section III-A. To introduce the semantic in-formation, we developed a customized automatic annotation framework to label MEMS LiDAR and 4D Radar point clouds in REHEARSE [9].

Our data annotation process has the following steps:

1) The unlabeled raw data is extracted from the dataset and stored in x, y, z, and intensity format, resulting in 143 sequences containing about 300 dense LiDAR point clouds per sequence.
2) We then estimate the road plane using the RANSAC algorithm and keep it for later usage.
3) We create bounding boxes for the rain sprinklers and label them. Points that are below the estimated road plane are omitted.
4) Next, we create bounding boxes for the other objects on the road surface (e.g., pedestrians, bicyclists, cars, target objects) and label them accordingly.
5) A 2D polygon bounding box for the road is then created.
6) Any points above the road plane, within the 2D polygon bounding box, that do not belong to the previously labeled classes are then labeled as raindrops.
7) The remaining points on the 2D polygon are further labeled as a part of the road.
8) All the remaining points are labeled as background.
9) Since the object bounding boxes may vary slightly from sequence to sequence due to noise in the sensor readings, we manually correct the bounding boxes for each sequence and repeat the process from steps 2 to 8.
10) The corresponding 4D Radar sparse point clouds are then automatically annotated by transferring labels from the nearest LiDAR points.

### C. Data Statistics

There are, in total, 143 sequences annotated in REHEARSE-3D. As illustrated in Figure 3, 39% of these sequences are obtained under clean conditions, whereas the remaining 61% involve rain emulated at varying densities.

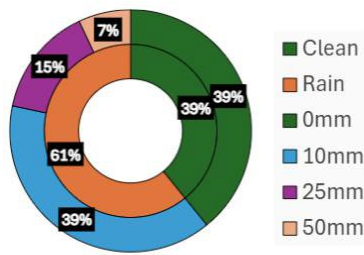

Fig. 3: REHEARSE-3D Data Distribution.

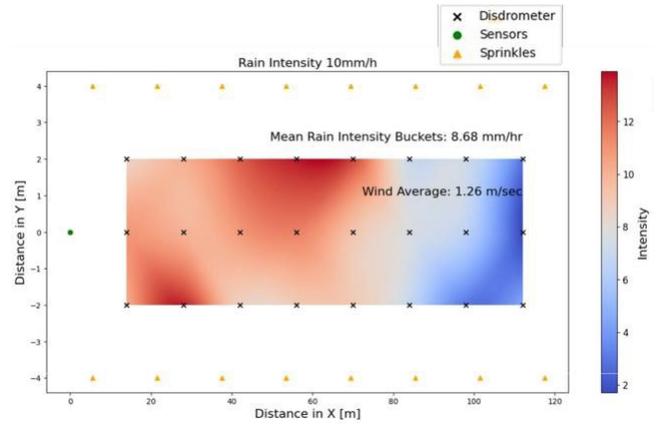

Fig. 4: Intensity heat map of the emulated rain at the CARISSMA outdoor test track – nominal intensity of 10 mm/h. Uniformity is influenced by wind.

Figure 4 shows the intensity heat map of the synthetic rain generated during outdoor testing at CARISSMA. The rain intensity is measured in each square meter for a nominal 10 mm/h rain intensity. In the heat map, the red region indicates high intensity, with values up to 14 mm/h, while the blue area corresponds to low intensity. As shown in Figure 4, the uniformity persists up to a distance of 80 meters from the sensor ring. Within this range, the intensity fluctuates between 8 and 14 mm/h. Beyond the range of 80 m, a noticeable decline in intensity is observed, with values falling below 2 mm/h, reaching insignificant intensity values. The maximum rain accumulation is observed around (25, -2) and (58, 2), which refer to (X, Y) positions in meters. Although the average rain intensities align with the nominal values, the non-uniform intensity distribution is caused by wind, resulting from outdoor testing.

In REHEARSE-3D, the annotation comprises point-wise annotations of 8 semantic classes: sprinkler, car, pedestrian, bike, targets, road, rain, and background. Figure 5 depicts the comprehensive statistics of the point-wise annotations. It is evident from the figure that certain classes (such as back-ground, road, and rain) predominate, while others, including pedestrians, bikes, and cars, exhibit a lower frequency of occurrence. It is noteworthy that this class imbalance problem is a pervasive issue in many existing datasets.

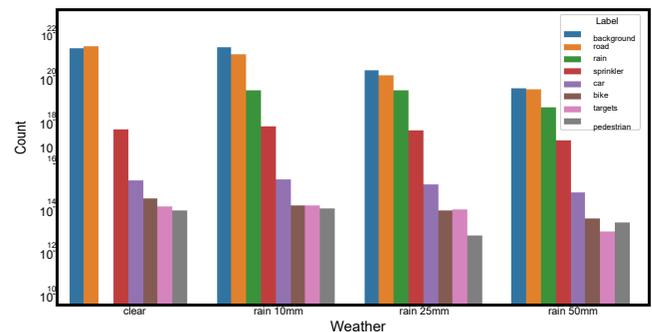

Fig. 5: Number of annotated points in each semantic class in REHEARSE-3D, separated by rain densities.

To the best of our knowledge, REHEARSE-3D is the first large-scale, fully point-wise annotated dataset to benchmark emulated rain data in 3D space. Table I provides a compara-tive analysis of REHEARSE-3D with several existing point cloud datasets.

D. Data Illustration

Figure 1 shows a sample annotated LiDAR and 4D Radar point cloud data with the corresponding RGB and thermal images. Note that these images are not annotated but are shown only for the sake of clarity.

Furthermore, Figure 6 shows examples of annotated 3D point cloud data captured in clean and rainy conditions in daytime and nighttime.

E. Rain Simulation

In addition to the rain-emulated sequences provided in REHEARSE-3D, we create corresponding rain-simulated counterparts for comparative evaluation. We adopt the physically accurate LiDAR rain model proposed by Espineira et al. [31] and apply it to clean point cloud sequences. Unlike Espineira et al.'s original implementation, which simulates each laser beam within a simulation engine (e.g., Unreal Engine), we apply the model to real-world recordings. To this end, we introduce a pre-processing step to retrieve each laser beam data from recorded point clouds, enabling the identification of unreturned beams. We project the point clouds onto Polar Grid Maps (PGMs) using the known azimuth and elevation angles of the calibrated LiDAR sensor. Unreturned beams are represented as points with maximum range and zero intensity, and the PGMs are subsequently reshaped back into list-based point cloud format, as detailed in Algorithm 1. Following the same logic as Espineira et al.'s

---

**Algorithm 1** Pre-processing to retrieve unreturned LiDAR beams. Inputs: $X \in R^{N \times 3}$ (3D Cartesian coordinates), $I \in R^N$ (intensity values), $\Theta_{calib} \in R^V$ and $\Phi_{calib} \in R^H$ (calibrated elevation and azimuth angles), $R_{max} \in R$ (maximum sensor range).

1: **function** PREPROCESS($X, I, \Theta_{calib}, \Phi_{calib}, R_{max}$)
2:    Init: $X_{pgm} \in R^{V \times H \times 3}$, $I_{pgm}, R_{pgm} \in R^{V \times H}$ as zeros
3:    $(R, \Phi, \Theta) \leftarrow \text{ToPolar}(X)$
4:    $\Theta_{idx} \leftarrow \text{argmin}_i |\text{sort}(\Theta_{calib})[i] - \Theta|$
5:    $\Phi_{idx} \leftarrow \text{argmin}_j |\text{sort}(\Phi_{calib})[j] - \Phi|$
6:    $X_{pgm}[\Theta_{idx}, \Phi_{idx}] \leftarrow X$; $I_{pgm}[\cdot] \leftarrow I$; $R_{pgm}[\cdot] \leftarrow R$
7:    $M \leftarrow (R_{pgm} == 0)$; $R_{pgm}[M] \leftarrow R_{max}$
8:    $\Phi_{full}, \Theta_{full} \leftarrow \text{meshgrid}(\Phi_{calib}, \Theta_{calib})$
9:    $X' \leftarrow \text{ToEuclidean}(R_{pgm}, \Phi_{full}, \Theta_{full})$
10:   $X_{pgm}[M] \leftarrow X'[M]$
11:   **return** $X_{pgm}.\text{reshape}(-1, 3)$, $I_{pgm}.\text{reshape}(-1)$
12: **end function**

---

model, if a raindrop is detected within the maximum range of an unreturned beam, the resulting point is added to the point cloud; otherwise, the beam remains unreturned.

Raindrop diameters range from 0.5 mm to 6 mm, and the rain is simulated at intensities of 10, 25, and 50 mm/hr, con-sistent with the rain-emulated settings in REHEARSE-3D. The resulting dataset is aligned with the original distribution of sequences, targets, and annotation classes to ensure a fair comparison.

We name this simulated dataset WMG Simulated data as the simulation was performed in Warwick Manufacturing Group (WMG) at the University of Warwick.

IV. BENCHMARK ON 3D POINT CLOUD DERAINING

In inclement weather conditions, the quality of the 3D point clouds captured by LiDAR and 4D Radar sensors can be substantially degraded by precipitation (e.g., raindrops). The harsher the precipitation is, the noisier the point cloud becomes. As shown in recent works, such degradation can lead to substantial drops in the performance of the downstream perception tasks, such as semantic segmentation [2–4] and object detection [4]. This challenge has prompted exten-sive research on detecting and removing noisy point cloud data points. Such point cloud denoising (a.k.a. deraining) is a binary segmentation task where each rain drop is treated as an outlier to be filtered out.

Leveraging REHEARSE-3D, we explore detecting and removing emulated raindrops from early merged LiDAR and 4D Radar point clouds. More specifically, we perform an in-depth empirical study showing the performance of several statistical and deep learning models on deraining REHEARSE-3D point clouds, including clean and rainy cases in both daytime and nighttime conditions, as shown in Figure 6.

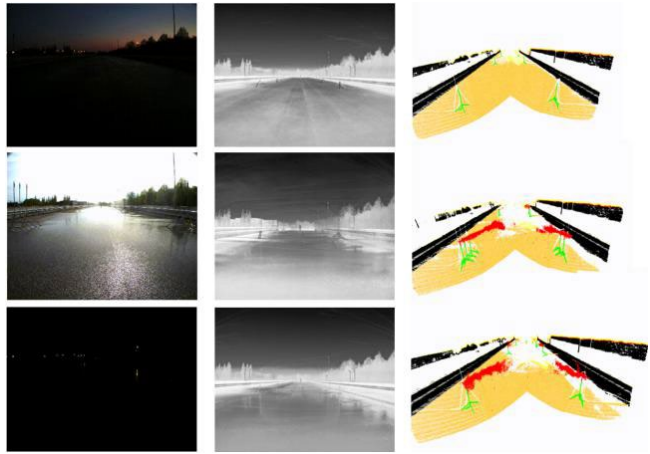

Fig. 6: Annotated sample LiDAR point clouds (right) to-gether with the corresponding RGB (left) and thermal camera images (middle) from nighttime-clean, daytime-rain, and nighttime-rain conditions (from top to bottom), respectively.

A. Benchmark Setup

Baselines: We employ some unsupervised statistical filtering approaches. Radius Outlier Removal [35] (ROR)

TABLE II: Rain drop detection results on the REHEARSE-3D validation and test splits. All scores are in percentages. S and US stand for Supervised and Unsupervised training, respectively. ↓ denotes lower is better, while ↑ indicates higher is better. Computation time scores are taken from [3].

| Model | Type | Validation | | | | Test | | | | Time (ms)↓ |
|---|---|---|---|---|---|---|---|---|---|---|
| | | Precision↑ | Recall↑ | $F_1$↑ | mIOU↑ | Precision↑ | Recall↑ | $F_1$↑ | mIOU↑ | |
| 3D-OutDet [3] | Deep Learning- S | 96.35 | 98.48 | 97.40 | 94.94 | 97.23 | 96.61 | 96.92 | 94.03 | 82 |
| SalsaNext [32] | Deep Learning- S | 95.17 | 99.41 | 97.24 | 94.61 | 95.81 | 99.07 | 97.41 | 94.92 | 97 |
| LiSnowNet-L1 [33] | Deep Learning - US | 14.42 | 31.94 | 19.55 | 12.44 | 21.22 | 37.70 | 26.07 | 16.86 | 131 |
| DSOR [8] | Statistical- US | 15.50 | 38.18 | 22.05 | 12.39 | 25.73 | 49.33 | 33.82 | 20.35 | 253 |
| DROR [34] | Statistical- US | 5.88 | 76.32 | 10.91 | 5.77 | 10.78 | 73.48 | 18.81 | 10.38 | 199 |

identifies a point as noise if it has fewer neighbors than a predefined threshold within a specified radius. Statistical Outlier Removal [35] (SOR) removes a point in case it lies further away from its neighbors than average. Dynamic ROR [34] (DROR) and Dynamic SOR [8] (DSOR) are dynamic extensions of ROR and SOR, respectively. DROR and DSOR are far superior than the ROR and SOR, hence we only run the denoising with DROR and DSOR. DROR differs from ROR in that it computes a dynamic radius by reviewing the density of distant points [34]. In DSOR, a local distance threshold is computed dynamically by considering the data statistics [8]. It should be noted that all these statistical methods are computationally expensive.

Additionally, we train three deep neural networks: SalsaNext [32], LiSnowNet-L1 [33], and 3D-OutDet [3]. SalsaNext [32] is a general-purpose semantic segmentation model trained to remove rain points as a binary-segmentation task. In SalsaNext [32], 3D point clouds are converted to 2D range view images to be further segmented by an advanced Convolutional encoder-decoder architecture. On the other hand, LiSnowNet [33] is introduced as a task-specific model to de-noise point clouds in an unsupervised fashion. LiSnowNet is specifically designed for point cloud data corrupted with snowfall. Therefore, it relies on the assumption that natural signals are sparse under the Fourier Form Transformation and the Discrete Wavelet Transformation [33]. However, when point clouds are corrupted with snowfall, the signal becomes less sparse under those transformations. The model 3D-OutDet [3] is a light-weight deep learning model that considers the local neighborhood relations to search for noisy points.

Evaluation Metric: Following the existing literature [3, 6], to assess the 3D point cloud draining performance we employ precision = $\frac{TP}{TP+FP}$, recall = $\frac{TP}{TP+FN}$, $F_1$ score, and mean Intersection over Union (mIoU) as quantitative metrics. Specifically, the $F_1$ score and mIoU are defined as follows:

$$F_1 = 2 \times \frac{\text{precision} \times \text{recall}}{\text{precision} + \text{recall}}, \quad (1)$$

$$IoU = \frac{TP}{TP+FP+FN}, \quad (2)$$

where $TP$ denotes True Positives as the correctly identified rain points, FP refers to False Positives as the non-rain points that are misidentified as rain, and FN indicates False Negatives as the instances where rain points were not detected correctly.

Implementation Details: We split REHEARSE-3D into three parts, including 25507 scans of full 3D data for training, 8301 for validation, and 8602 for testing. Train and test splits are balanced with a similar proportion of point clouds from clean and rainy conditions.

We follow the training protocol in [3] and report the model with the best performance. The statistical filters are tuned using Optuna[1]. Parameter search is done on a subset of 100 random samples for 100 iterations.

B. Benchmark Results

Quantitative Results: Table II shows the de-raining scores obtained on REHEARSE-3D for different state-of-the-art deep learning and statistical methods. Note that with the exception of SalsaNext [32] and 3D-OutDet [3], all methods are trained in an unsupervised manner.

3D-OutDet [3] has the best precision and the lowest computational cost, while SalsaNext [32] has the best recall for both the validation and test sets. Again, 3D-OutDet [3] has the best $F_1$ score and IOU for the validation set, whereas SalsaNext [32] has the best $F_1$ score and IOU for the test set. However, we obtain a significant performance drop in all the unsupervised models. This clearly shows that a network trained with well-annotated supervision signals can accurately learn the underlying distribution of the emulated raindrops.

It is noteworthy that LiSnowNet-L1 [33] performs relatively poorly in detecting raindrops. We strongly believe that this is because LiSnowNet-L1 [33] is explicitly designed as a task-specific model to detect snowflakes only. The model is carefully designed to identify sparsity in point clouds subjected to specific transformations, such as the Fast Fourier or Discrete Wavelet transformations. However, these hard constraints do not pertain to raindrops, yielding the lowest detection scores.

Table III shows the de-raining quantitative results for both WMG Simulated data and the REHEARSE-3D emulated data. The experiments are separated according to rain intensity (heavy, medium, and light). In this experiment, we employ a supervised deep learning model (3D-OutDet [3]) and an unsupervised statistical model (DSOR [8]) for bench-marking. We notice that both models perform better on the WMG simulated data, and the deep learning based 3D-OutDet [3] has almost 100% accuracy in all rain inten-sities. This is because this simulation is generated as per Algorithm 1, and the 3D-OutDet [3] can easily capture the simulation process. Note that the same model performs less on the emulated REHEARSE-3D dataset, which is not mathematically generated. Even though the rain has been emulated using water sprinklers, the environmental factors (e.g., wind direction, wind velocity, gravity, temperature, etc) have shifted the emulated rain data distribution towards real rain data distribution. As a result, both 3D-OutDet [3] and DSOR [8] face challenges in de-raining the emulated data in REHEARSE-3D.

---
[1] https://optuna.org/

TABLE III: Performance Comparison between simulated (WMG simulated) and emulated (REHEARSE-3D) data. ↓ denotes lower is better, while ↑ indicates higher is better.

| Model | Rain Density | WMG Simulated | | | | REHEARSE-3D | | | |
|---|---|---|---|---|---|---|---|---|---|
| | | Precision↑ | Recall↑ | F1↑ | mIOU↑ | Precision↑ | Recall↑ | F1↑ | mIOU↑ |
| 3D-OutDet | Heavy | 99.99 | 99.99 | 99.99 | 99.99 | 98.30 | 96.43 | 97.35 | 94.85 |
| | Medium | 99.99 | 99.99 | 99.99 | 99.99 | 97.88 | 95.89 | 96.87 | 93.93 |
| | Light | 99.99 | 99.99 | 99.99 | 99.99 | 98.08 | 98.67 | 98.38 | 96.81 |
| DSOR | Heavy | 78.58 | 99.59 | 87.85 | 78.33 | 27.08 | 39.46 | 32.12 | 19.13 |
| | Medium | 69.00 | 99.81 | 91.59 | 68.91 | 38.24 | 56.27 | 45.54 | 29.48 |
| | Light | 55.03 | 94.34 | 70.97 | 55.00 | 23.35 | 35.95 | 28.31 | 16.49 |

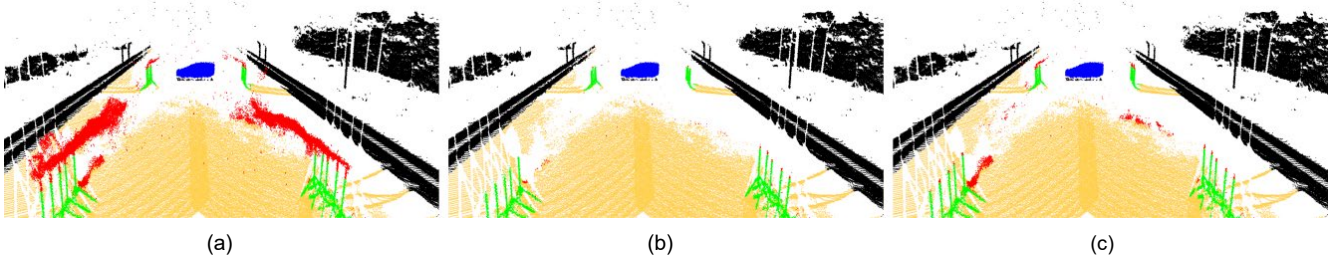

(a)          (b)          (c)

Fig. 7: (a) Original emulated rainy scene, (b) Rain removed by SalsaNext [32], and (c) Rain removed by 3D-OutDet [3]

Computation Time: For the computation time, we report the total execution time provided in [3], which includes all pre-processing, post-processing, and runtime scores in milliseconds. As reported in [3], the runtime experiments were conducted on an Ubuntu 20.04 system featuring dual Intel ®Xeon(R) Silver 4210R CPUs, two Nvidia RTX A6000 GPUs, and 192 GB of system memory.

In our experiments, SalsaNext [32] stands out as the best model for deraining 3D point clouds, whereas 3D-OutDet [3] is the most economical choice, offering the lowest latency.

Qualitative Results: Figure 7 depicts sample qualitative results for SalsaNext [32] and 3D-OutDet [3]. This figure clearly shows that these models can, to a great extent, detect and filter out raindrops while preserving other irrelevant points.

V. DISCUSSION & CONCLUSION

In this paper, we introduced REHEARSE-3D as a new multimodal dataset for weather-aware autonomous driving research, offering a comprehensive collection of emulated rain data. As the first dataset of its kind, REHEARSE-3D features point-wise semantically annotated high-resolution LiDAR-256 scans coupled with 4D Radar point clouds, captured in both daytime and nighttime rainy conditions. REHEARSE-3D goes beyond typical datasets by providing detailed precipitation characteristics, including rain intensity, droplet size distribution, wind data, and visibility informa-tion. This wealth of data enables not only sensor noise modeling but also the point-level analysis of weather impacts on sensors.

We demonstrate the utility of our REHEARSE-3D dataset as a novel point cloud de-raining benchmark by comparing the performance of various statistical and deep learning models on REHEARSE-3D. Our findings in Table II clearly show that REHEARSE-3D can serve as a novel test bed for evaluating and comparing advanced future de-raining approaches.

Limitations: REHEARSE-3D involves only front-view static scenes in a controlled environment; therefore, neither the sensor ring nor objects in the scene (e.g., cars, pedestri-ans, etc.) are moving. As in the case of all other emulated rain data, there still exists a gap between our generated and real rain data. This gap is, however, less than that observed between the simulated and real data. Though there are RGB and Thermal camera images and Ouster LiDAR data are available, we only annotated the MEMS LiDAR point clouds merged with 4D Radar data.